\documentclass{article}

\usepackage[final,nonatbib]{neurips_ai4d3_2023}

\usepackage[utf8]{inputenc} % allow utf-8 input
\usepackage[T1]{fontenc}    % use 8-bit T1 fonts
\usepackage{hyperref}       % hyperlinks
\usepackage{url}            % simple URL typesetting
\usepackage{booktabs}       % professional-quality tables
\usepackage{amsfonts}       % blackboard math symbols
\usepackage{nicefrac}       % compact symbols for 1/2, etc.
\usepackage{microtype}      % microtypography
\usepackage{xcolor}         % colors
\usepackage{graphicx}
\usepackage{subcaption}
\usepackage{amsmath}

\usepackage[numbers]{natbib}

\title{GraphPrint: Extracting Features from 3D Protein Structure for Drug Target Affinity Prediction}

\author{%
  Amritpal Singh \\
  Department of Computer Science\\
  Georgia Institute of Technology, USA\\
  \texttt{asingh880@gatech.edu} \\
}

\begin{document}

\maketitle
% \begin{abstract}
% TODO
% \end{abstract}

% \begin{IEEEkeywords}
% Drug-Target Affinity, GNN
% \end{IEEEkeywords}

%%%%%%%%% ABSTRACT
\begin{abstract}
Accurate drug target affinity prediction can improve drug candidate selection, accelerate the drug discovery process, and reduce drug production costs. Previous work focused on traditional fingerprints or used features extracted based on the amino acid sequence in the protein, ignoring its 3D structure which affects its binding affinity. %Recent works like alphafold have improved access to 3d protein structure prediction.
In this work, we propose GraphPrint: a framework for incorporating 3D protein structure features for drug target affinity prediction. We generate graph representations for protein 3D structures using amino acid residue location coordinates and combine them with drug graph representation and traditional features to jointly learn drug target affinity. Our model achieves a mean square error of 0.1378 and a concordance index of 0.8929 on the KIBA dataset and improves over using traditional protein features alone. Our ablation study shows that the 3D protein structure-based features provide information complementary to traditional features. 
\end{abstract}

% === INTRODUCTION ===
\section{Introduction}
% Answer: What problem are you trying to solve or investigate?
Predicting drug and target affinity (DTA) can improve drug candidate selection, accelerate drug discovery duration, and reduce drug production costs, repurposing existing drugs \citep{DeNovo}. Different techniques such as molecular coupling \citep{Docking1,Docking2,Docking3}, similarity-based methods, \citep{KronRLS,SimBoost}, network-based methods \citep{Modality}, and deep learning-based methods. Deep learning methods have been used to perform DTA, and have gained popularity in recent times as they can have higher accuracy and lower prediction time. \citep{DeepDTA,GraphDTA,GDGRU,Affinity2Vec,iEdge} 

\begin{figure*}[htbp]
\centering
\includegraphics[width=0.6\linewidth]{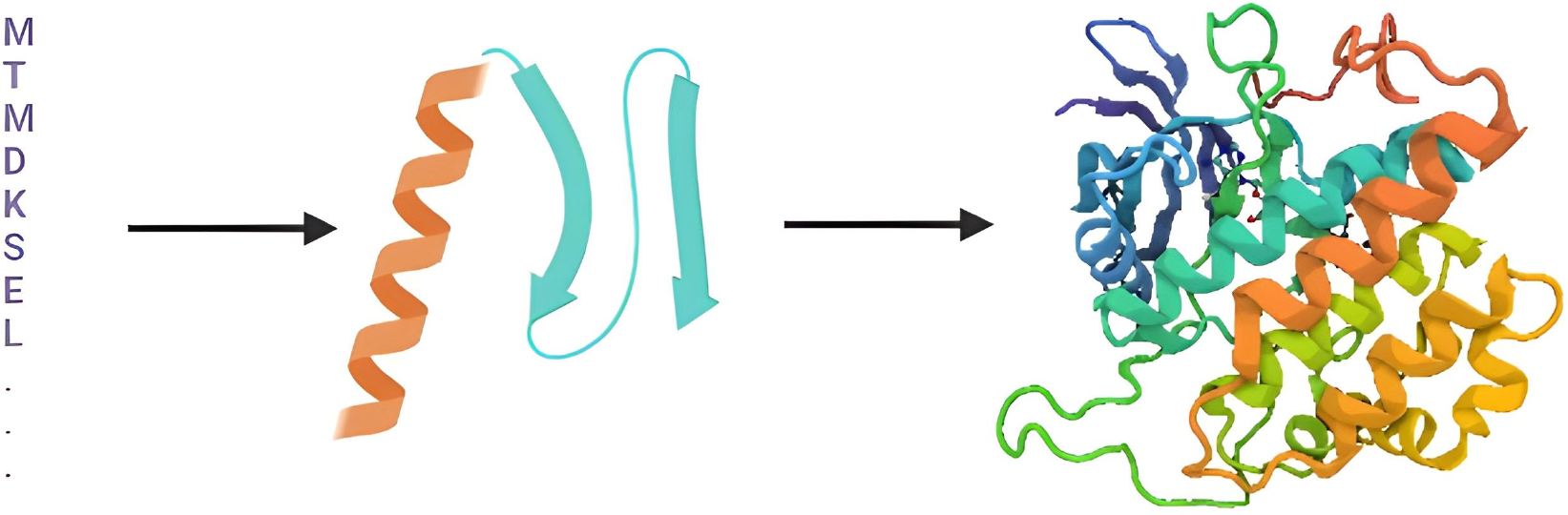}    
\caption{Protein structure can be visualized at several levels. The primary structure involves the structure of amino acids in proteins. The secondary structure involves the formation of helix and sheet structures. The tertiary structure involves the folding of the amino acid chain into a 3D space. Proteins with more than one peptide chain can have a quaternary structure, which involves further folding of chains over each other. }
\label{fig:prot_structure_levels}
\end{figure*}

% \citep{DeepDTA,GraphDTA,GDGRU,DGraphDTA,Affinity2Vec,DeepCPI,S2GC,SMT,SEGSA,iEdge,DeepGS}
Existing deep learning-based methods use amino acid residue sequences in protein to extract features directly or learn feature extractors. \citep{DeepDTA,GraphDTA,GDGRU,Affinity2Vec,iEdge}. The amino acid sequence can be of primary, secondary, tertiary, and sometimes quaternary structure, as shown in figure \ref{fig:prot_structure_levels}. The primary structure consists of a sequence of amino acids, and the secondary structure represents the alpha-helical or beta-sheet structure. The tertiary structure consists of folding the chain itself, leading to a 3D structure. Proteins with more than one peptide sequence can have additional folding among each other, leading to a quaternary structure. These higher structures affect the bindings and docking sites of these proteins. Incorporating the 3D structure of a protein can help improve performance on tasks such as drug target affinity prediction. 
With the advent of accurate 3D protein structure prediction \citep{alphafold}, it is now possible to
predict the 3D structure of proteins based on amino acid sequence. With the growing size of the 3D protein structure database, incorporating 3D features to learn drug-related features can be a new direction to explore in the field of drug discovery.

DTA has been framed as a binary classification and regression problem. In binary classification, binary labels 0 and 1 are predicted. In a regression problem, drug-target affinity is quantified as a regression task, which can be used to rank drug targets in order of their binding affinity. This can be important for selecting only a limited number of drug candidates for further investigation in the lab. 

In this work, we propose a graph neural network-based architecture to learn features based on the 3D structure of proteins. More specifically, our contributions are:

\begin{enumerate}
    \item We propose GraphPrint, a framework to integrate 3D protein structural representation to learn features using graph neural networks.
    \item We perform an ablation study to show that the 3D structure of protein provides complementary information to traditional handcrafted features.
    \item We share a curated version of the KIBA dataset along with its 3D protein structure to help future works.
\end{enumerate}

\section{Related works}
Different techniques such as molecular coupling \citep{Docking1,Docking2,Docking3}, similarity-based methods, \citep{KronRLS,SimBoost}, network-based methods \citep{Modality}, and deep learning-based methods. Deep learning methods have increasingly been used for drug target affinity prediction.
Table \ref{tab:previous_work} shows some of the previous state-of-the-art models and their backbone architecture. DeepDTA\citep{DeepDTA} was one of the first deep learning-based architectures for DTA tasks. The authors used 1D convolution to embed the drug’s SMILES representation and protein sequence separately, followed by concatenation and classifier training.

To improve feature extraction, graph neural network-based architectures have been proposed. GraphDTA\citep{GraphDTA} is one such earlier work, to use graph neural networks to learn a drug structure representation, combined with CNN for protein sequence. iEdgeDTA \citep{} treats protein sequence as a 1d graph and uses graph convolutional layers to extract protein features, combined with edge-graph convolutional for drug embeddings. BiCompDTA use normalized compression distance and Smith-Waterman measures to generate protein embeddings, that is later used to train deep learning network. Table \ref{tab:previous_work} summarizes the architecture and input data representations used in these methods.

Most previous works treat protein as a sequence of amino acids to either extract hand-crafted features or learn embeddings using neural network-based architectures. This work ignores the secondary, tertiary, or quaternary structure of proteins. These higher structures affect the bindings and docking sites of these proteins. Incorporating the 3D structure of a protein can help improve performance on tasks such as drug target affinity prediction. With recent works like Alphafold \citep{alphafold}, it is now possible to predict 3D structures of proteins using their aminoacid sequence.

\begin{figure*}[htbp]
\centering
\includegraphics[width=\linewidth]{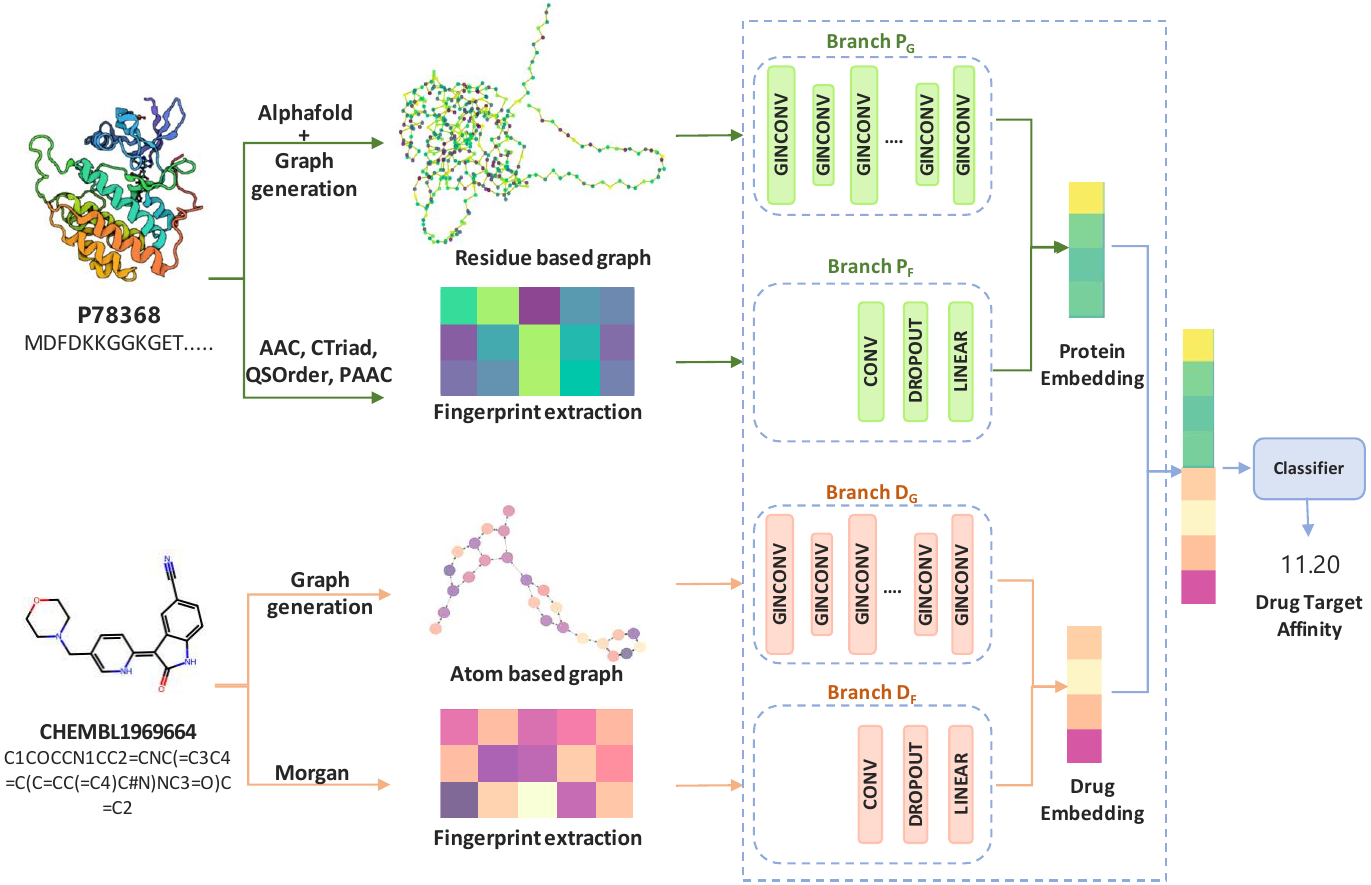}    
\caption{Diagram showing pipeline for feature extraction for protein using Aplhafold \citep{alphafold} and architecture for GraphPrint, with multihead architecture containing graph isomorphic convolution layers (GINCONV) and 1D convolutional blocks, followed by concatenation of features into a multilayer perceptron as a classifier.}
\label{fig:architecture}
\end{figure*}

\begin{table*}[ht]
\centering
\caption{Previous work for drug target affinity prediction, along with their architectures and inputs used. Keywords: NCD= Normalised compression distance.
}
\label{tab:previous_work}
\begin{tabular}{llllll}
\hline
  &   & \multicolumn{2}{c}{\textbf{Protein}} & \multicolumn{2}{c}{\textbf{Drug}} \\
\multicolumn{1}{c}{\textbf{Model}} &
  \multicolumn{1}{c}{\textbf{Year}} &
  \multicolumn{1}{c}{\textbf{Input}} &
  \multicolumn{1}{c}{\textbf{Backbone}} &
  \multicolumn{1}{c}{\textbf{Input}} &
  \multicolumn{1}{c}{\textbf{Backbone}}  \\
\hline
\\
DeepDTA        & 2018          & AA sequence         & CNN             & SMILES            & CNN           \\
GraphDTA       & 2020          & AA sequence         & CNN             & SMILES            & GNN           \\
iEdgeDTA       & 2023          & AA sequence         & 1D-GCN          & SMILES            & Edge-GCN      \\
BiComp-DTA     & 2023          & AA sequence & NCD features               & SMILES            & CNN            \\
\hline
\end{tabular}
\end{table*}

% === METHODS ===
\section{Methods}
% Answer: What's new in your study? 
%   Why does this matter, what are the implications? 
%   What methods will you be using? 
%   What metrics will you use to evaluate the success of the study?

In this section, we describe a framework for a graph representation of protein 3D structure and our architecture, followed by the data set details. 

\subsection{Architecture}
% Drug encoding branch, Protein encoding branch, Graph branch
We explore a multimodality approach by using a multi-head network to learn protein/drug embeddings directly from their structure as well as their extracted fingerprints. This allows the model to leverage the complementary information in these representations. 

We create a multihead neural network consisting of four branches. We use 
a graph convolution-based network for graph representations of proteins/drugs. 
For fingerprints, we use 1D convolutional blocks to learn embeddings. Figure \ref{fig:architecture} shows a pictorial representation of the architecture used. 

Mathematically, for any drug molecule $D_i$ and protein sequence $P_i$, we learn a model $F_c$, as shown in equation \ref{model_equation}, where $F_{P_{G}}$, $F_{P_F}$, $F_{D_G}$, $F_{D_F}$ represent branches $P_G$, $P_F$, $D_G$, $D_F$ of the architecture, respectively. The output from three branches is concatenated, before passing through the classifier $F_c$, generating a final prediction of the affinity score $y_i$.

\begin{equation}
\centering
    y_i = F_c([concat ( F_{P_G}(P_i), F_{P_F}(P_i), F_{D_G}(D_i), F_{D_F}(D_i) ])
% \caption{Architecture}
\label{model_equation}
\end{equation}

Branch $P_G$ and $D_G$ consist of 5 graph blocks in series, followed by a global pool average and a linear block. This helps to extract graph-level feature embeddings from the drug molecule. We add bottlenecks in the architecture to improve the information transfer enforcing model to learn efficient information on embedding information on embedding and lower parameter size. 
The branches $P_F$ and $D_F$ consist of a 1D convolutional layer and a linear layer. Outputs from all branches are concatenated and passed through fully connected layers to generate the final affinity score.

\begin{figure*}[htbp]
\centering
\includegraphics[width=\linewidth]{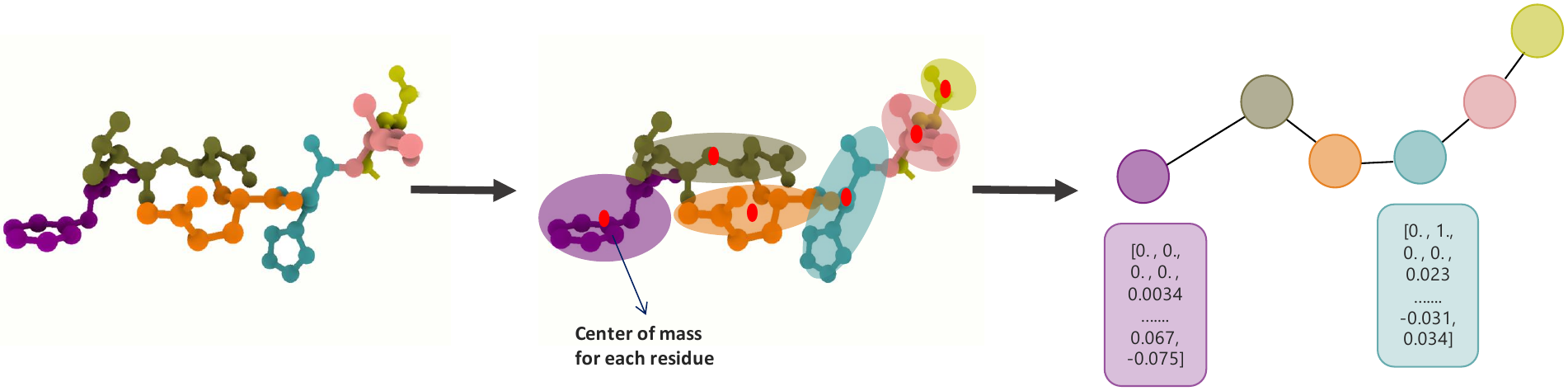}    
\caption{Protein graph representation: We calculate the center of mass of the amino acid and use this as a center of mass for the amino acid residue. Each residue represents a node, with node features containing position and amino acid properties.}
\label{fig:graph_creation}
\end{figure*}

\subsubsection{Graph embeedings}
To generate protein graph representations, we use Alphafold \citep{alphafold} to generate the 3D protein structure of each protein target. Using the 3D structure, we define each amino acid residue as a node. We use the residue's coordinates of the center of mass to represent amino acids. We add additional features, such as amino acid encoding, molecular weight, polarity, solubility, and pKa values, to each node feature. Figure \ref{fig:graph_creation} shows the creation of graph representation for protein structure. 
For drugs, we convert the SMILES string into a molecular graph representation using the open-source library RDKit\citep{RDKit}. Atoms are represented as nodes, and the bonds between them are represented as edges. For node embeddings, we use one-hot encodings of atom type, number of neighboring atoms, number of neighboring hydrogens (explicit and implicit), and implicit valence and aromaticity of molecule.

\subsubsection{Fingerprint extraction}
We extract traditional hand-designed features to augment the graph representation. We choose AAC, Conjoint triad fingerprint, and quasi-sequence fingerprint, as they only require the protein's amino acid sequence to acquire. AAC encodes \textit{k}-mers of amino acids into an 8,420-length bit vector. It can contain sequence neighborhood (local) information. The conjoint triad fingerprint utilizes a hand-made 7-letter alphabet to encode a continuous frequency distribution of three amino acids. This transforms the protein into a homogeneous vector space of a 343-length vector. The quasi-sequence fingerprint contains a residue pair correlation within its 100-length vector. All three of these representations are concatenated before being passed to the network. We use the open-source library iFeature \citep{iFeature} to calculate protein representations.   

For drug molecules, we extract Morgan and Daylight fingerprints using the drug SMILES sequence, generating output of size 1024 and 2048 respectively. Morgan's fingerprint encodes circular radius-2 substructures of a molecule with a 1024-length bit vector. This includes partially disambiguated atom identifiers. The daylight fingerprint encodes path-based substructures into a 2048-length bit vector. We use the Therapeutics Data Commons (TDC) open-source library\citep{TDC} to calculate these features.

% Datasets (Davis & KIBA)
\subsection{Datasets}
% Answer: What dataset(s) will you be using? Provide a brief description and necessary steps to process it if needed
In this work, we use the KIBA dataset for evaluation \citep{KIBA}. The KIBA dataset contains an affinity score that combines $K_d$, an inhibitor constant ($K_i$), and the half maximum inhibition concentration ($IC_{50}$). For comparison with previous works, we use a similar dataset generated by filtering all the drug-protein combinations with fewer than 10 interactions. The final dataset contains a total of 2,111 drugs and 229 proteins. The KIBA score ranges from 0.0 to 17.2 and a larger score represents a weaker binding affinity.

%   What metrics will you use to evaluate the success of the study?
\subsection{Evaluation}
After model training is complete, we freeze the architecture and measure the performance on the test set.  We will use the following metrics: Mean Squared Error (MSE) as defined as $MSE = \frac{1}{n}\sum_{i=1}^n (P_i - Y_i)^2 $, where $n$ is the number of data points, $\boldsymbol{P}$ are the predicted affinity values, and $\boldsymbol{Y}$ are the expected affinity values. Lower MSE is better. Concordance Index (CI): \citep{DeepDTA} compares the predicted order of the binding affinity values corresponding to drug-target interactions with ground truth. CI values greater than 0.8 indicate a strong model. It is defined as $CI = \frac{1}{Z}\sum_{y_i>y_j} h(p_i - p_j)$, where $h(x) = \begin{cases}1 \& x > 0, 0.5 \& x = 0, 0 \& x < 0 \end{cases}$. $r_m^2$ metric: measures external prediction performance of Quantitative structure-activity relationship (QSAR) models. This metric is defined as $r_m^2 = r^2 * \bigg(1 - \sqrt{r^2 - r_0^2}\bigg)$, where $r^2$ is the squared correlation coefficient with intercept, and $r_0^2$ is without intercept. A value above 0.5 is good. Spearman correlation measures if two variables are monotonically related, as defined by Spearman = $1 - \frac{6\sum (P_i - Y_i)^2}{n(n^2 - 1)}$. Pearson Correlation: a measure of the linear correlation between two variables, as defined by:  $Pearson = \frac{\mathrm{cov}(P,Y)}{\sigma_P \sigma_Y}$.

We explore the ablation of branches to understand their contributions to the learning of our architecture. We remove one layer at a time and report the metrics of the modified architecture. We also explore errors contributed by individual drugs and proteins to the overall error, by plotting the sum of MAE error per drug or protein.

\section{Results and Discussion}
In this study, we evaluate our trained model on the test subset of the KIBA dataset and compare it to previously reported baselines. Table \ref{tab:kiba_results} shows a comparison of the performance metrics of our models with DeepDTa, GraphDTA, iEdgeDTA, and BiCompDTA. Our model achieved 0.3713 RMSE,  0.1378 MSE, 0.8929 CI, Spearman correlation of 0.8852, and Pearson correlation of 0.8920. Our model results are competitive against state-of-the-art models. 

\begin{table*}[h]
\centering
\caption{Performance metrics our architecture with comparison to previous work on KIBA test dataset. Keywords: RMSE=root mean square, MSE= mean square error, CI=conformance index}

\label{tab:kiba_results}
\begin{tabular}{lllllll}
\hline
\textbf{Model}         & \textbf{RMSE}   & \textbf{MSE}            & \textbf{CI}             & \textbf{Spearman} & \textbf{Pearson} & \textbf{Epochs} \\
\hline
DeepDTA      & —      & 0.194          & 0.863          & —        & —       & -   \\
% Affinity2Vec      & —      & 0.124 & 0.91          & —        & —       & 1000   \\
GraphDTA      & —      &0.139 & 0.891          & —        & —       & 1000   \\
iEdgeDTA      & —      & 0.139 & 0.89           & —        & —       & -   \\
BiComp-DTA       & —   &  0.167 & 0.891         & —        & —       & -   \\
% ModalityDTA   & —      & \textbf{0.118} & \textbf{0.917} & —        & —       & —      
Ours ($P_G$,$P_F$,$D_G$)    &0.3926 & 0.1542 & 0.8790 & 0.8632 & 0.8748 &  300\\
Ours ($P_G$,$P_F$,$D_G$,$D_F$)  & 0.3713 & \textbf{0.1378 }& \textbf{0.8929} & 0.8852 & 0.892 &300  \\
\hline
\end{tabular}
\end{table*}

\begin{figure}[h]
  \centering
  \begin{subfigure}{0.45\textwidth}
    \includegraphics[width=200pt,height=150pt]{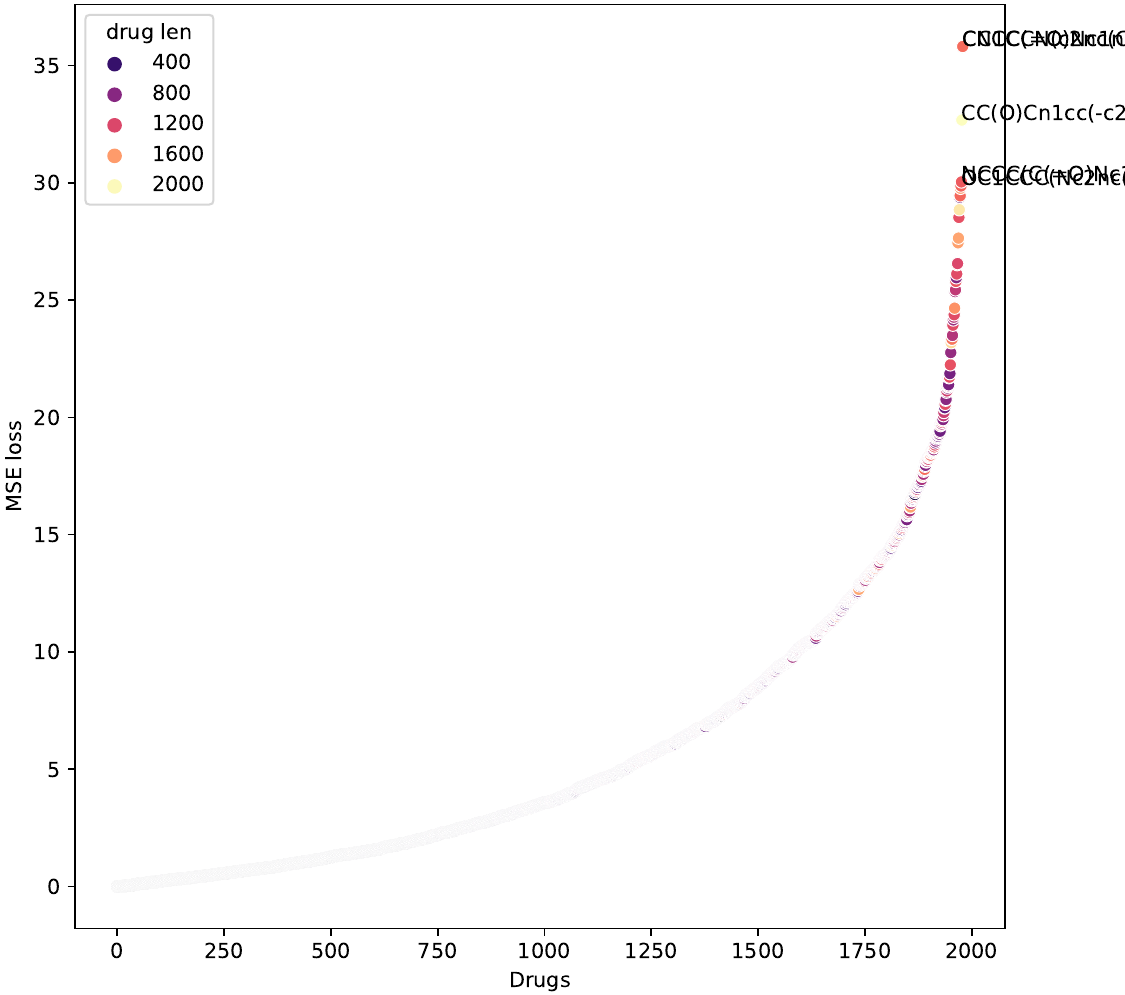}    
    \caption{Drug ID vs mse contribution}
    \label{fig:kiba_drug_error}
  \end{subfigure}
  \hfill
  \begin{subfigure}{0.45\textwidth}
    \includegraphics[width=200pt,height=150pt]{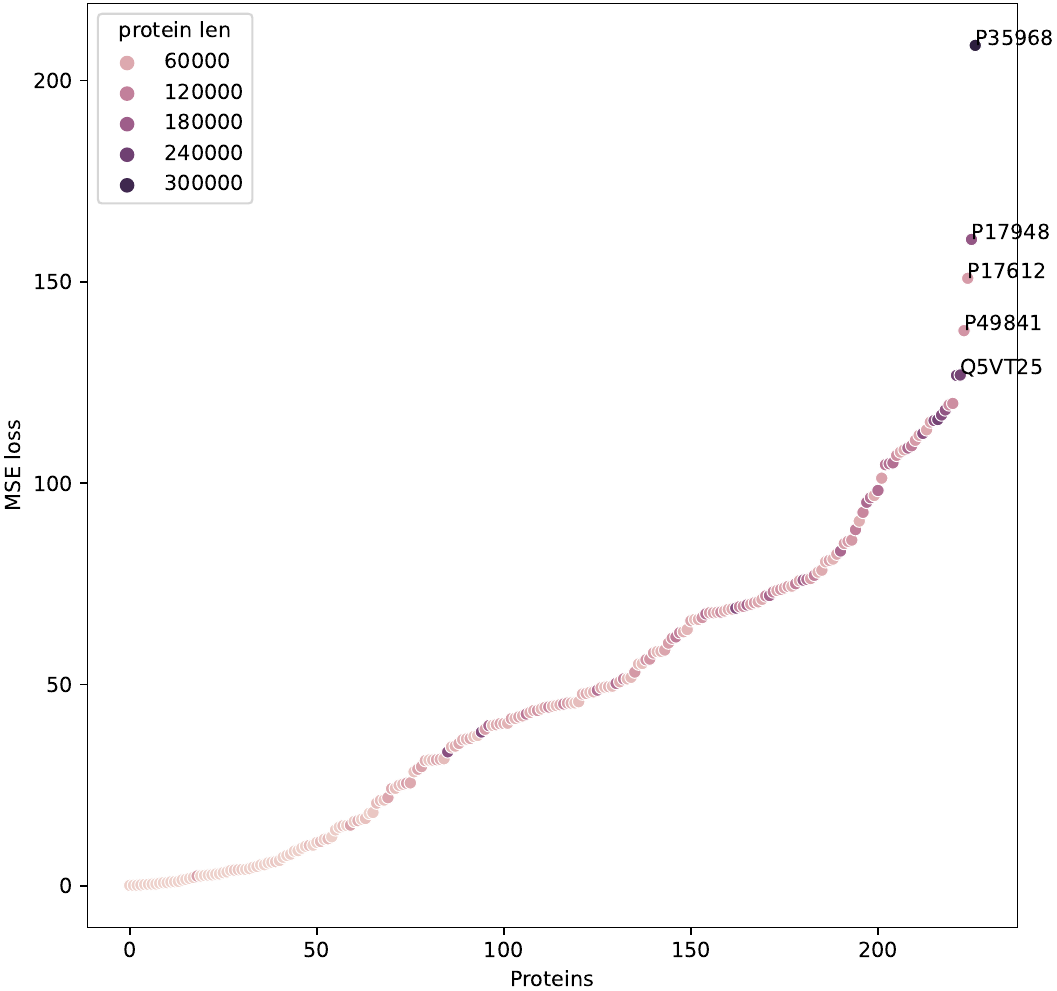}    
    \caption{Protein ID vs mse contribution}
    \label{fig:kiba_protein_error}
  \end{subfigure}
 \caption{Error breakdown based on drug ID, protein ID, aromatic compounds in drugs, bonds inside drug. A small amount of drugs and proteins contribute the most amount of error.}
  \label{fig:error_breakdown}
\end{figure}

\begin{figure}[h] %tbp]
  \centering
  \begin{subfigure}{0.3\textwidth}
    \includegraphics[width=110pt]{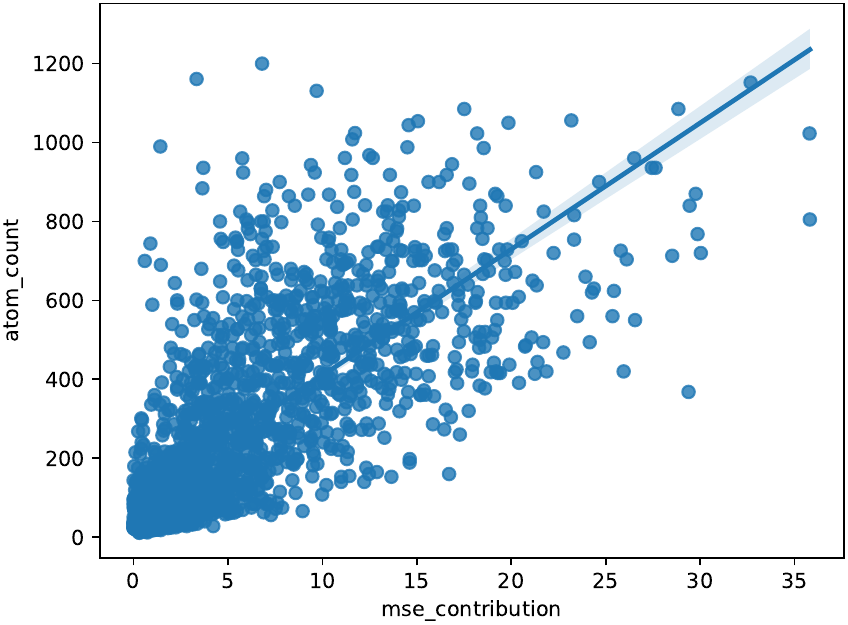}    
    \caption{Drug atom count vs mse contribution}
    \label{fig:KIBA_atom_count}
  \end{subfigure}
\hfill
  \begin{subfigure}{0.3\textwidth}
    \includegraphics[width=110pt]{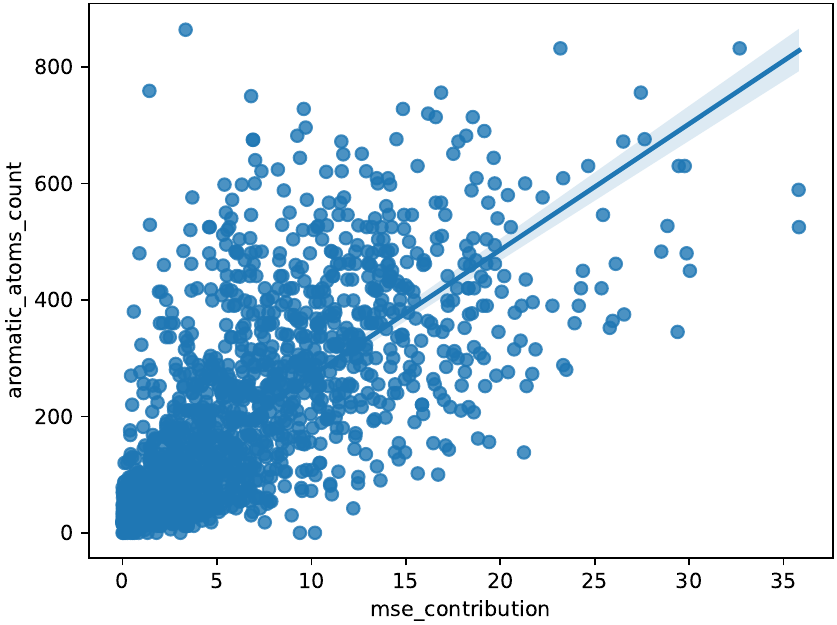}    
    \caption{Drug aromatic atoms count vs mse contribution}
    \label{fig:KIBA_arromatic_contribution}
  \end{subfigure}
\hfill
  \begin{subfigure}{0.3\textwidth}
    \includegraphics[width=110pt]{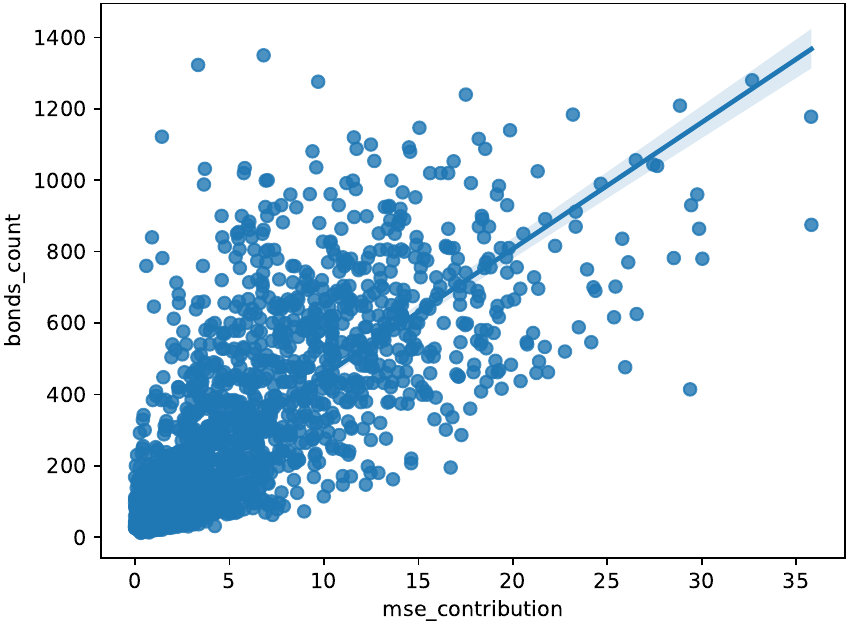}    
    \caption{Drug bonds count vs mse contribution}
    \label{fig:KIBA_bonds_count}
  \end{subfigure}
  
  \caption{Scatter plots showing the mse error contribution vs. different parameters. There is a linear relation between the number of atoms, aromatic atoms, and bonds to the error contribution of the respective drug molecule. }
  \label{fig:combined}
\end{figure}

We look into the MAE error contribution of individual proteins and drugs to the overall model MAE error. Figure \ref{fig:kiba_drug_error} plots the sum of errors for each protein and drug molecule. As shown in the figure, only a small number of protein and drug molecules are responsible for the majority of errors. Figure \ref{fig:KIBA_atom_count}, \ref{fig:KIBA_arromatic_contribution}, \ref{fig:KIBA_bonds_count} show scatterplot for the sum of error contribution per drug vs a number of drugs atom counts, aromatic atoms and bonds respectively. We can see that there is a linear relation with increasing atom count making drug affinity prediction harder. The same goes for the number of aromatic and number of bonds. 

\subsection{Ablation study}
We perform architecture ablations by removing one or more branches of the network. At each stage, the model receives one of the two embeddings for both protein and drug. Table \ref{tab:ablations} shows ablation results for our architecture. All the ablations have lower performance compared to our main architecture. Removing branch $P_G$ causes a dip in the concordance index and an increase in the MSE score. This supports our hypothesis that 3D structure provides complementary information to traditional hand-crafted fingerprints. Further, replacing simple architecture with a bottleneck architecture causes an increase of 1.3\% in the concordance index.

\begin{table*}[h]
\centering
\caption{Ablations results on different branches of our architecture on KIBA Dataset. Removing branch $P_G$ causes a dip in the concordance index, suggesting the complementary nature of 3D structure to traditional hand-crafted fingerprints.}
\label{tab:ablations}
\begin{tabular}{lllllll}
\hline
  \textbf{\textbf{Removed Branch}} &
  \multicolumn{1}{c}{\textbf{CI}} &
  % \multicolumn{1}{c}{\textbf{$\Delta$ CI}} &
  \multicolumn{1}{c}{\textbf{RMSE}} &
  \multicolumn{1}{c}{\textbf{MSE}} &
  \multicolumn{1}{c}{\textbf{Spearman}} &
  \multicolumn{1}{c}{\textbf{Pearson}} \\ \hline
- & \textbf{0.8929} &   \textbf{0.3713} & 0.1378 & 0.8852 & \textbf{0.892}  \\ 
$P_G$ & 0.8911 &   0.3740 & 0.1399 & 0.8824 & 0.8899 \\
$D_G$ & 0.8902 & 0.3716 & \textbf{0.1381} & \textbf{0.8924} & 0.8825  \\
$P_F$ & 0.8820 &   0.3857 & 0.1488 & 0.8838 & 0.8746   \\
$D_F$ & 0.8790 &   0.3926 & 0.1542 & 0.8632 & 0.8748 \\
$D_F$, $P_F$ & 0.8783 &   0.3938 & 0.1551 & 0.8626 & 0.8736 \\
$D_F$, $P_G$ & 0.8699 &   0.4104 & 0.168 & 0.8549 & 0.868 \\
\hline
\end{tabular}
\end{table*}

\subsection{Limitations and Future work}
There are a few limitations to improve on. Since the generation of 3D structures is a computationally expensive process, we present our results only on the KIBA dataset. More evaluation on multiple datasets is required for a more comprehensive evaluation. In this work, we focus on integrating 3D protein structure representation, we do not perform extensively on graph neural architecture. Using more recent architecture with attention can provide further boost performance. 
In the future, this work can be further extended and directly implement an explainability layer, pointing to amino acid regions interacting with drugs. It would be worthwhile to quantify error-contributing correlations in protein structures, as understanding the reasoning behind this might also lead to the development of better model architectures. 

\section{Conclusion}
In conclusion, our GraphPrint framework is a novel approach for drug target affinity prediction that leverages 3D protein structure features in addition to traditional protein and drug representations. We have demonstrated the potential of our model in improving drug candidate selection, accelerating drug discovery, and reducing production costs. By incorporating protein 3D structure information, our model achieved a mean square error of 0.1378 and a concordance index of 0.8929 on the KIBA dataset, surpassing the performance of models that rely solely on traditional protein features.

With the advent of accurate 3D protein structure prediction and increasing 3D protein databases, incorporating 3D protein structure for drug target affinity and other approaches can be a new direction to explore in the field of drug discovery and multimodal learning. This research not only enhances our understanding of drug-target interactions but also holds promise for more efficient and cost-effective drug development processes in the future. Further exploration and refinement of our model may pave the way for even more accurate predictions in this critical area of pharmaceutical research.

% \section{Code and dataset availability}
% To honor the blinded peer review, code and curated dataset of 3D protein structure will be released after the peer review process. code link: 

\section{Acknowledgement}
The author extends their gratitude to Nicholas Frato for his help in initial data processing and drug fingerprint integration for the pipeline.
% The author extends their gratitude to Nicholas Frato for his essential role in data processing and drug fingerprint integration for the pipeline.

% === REFERENCES ===


\begin{thebibliography}{00}
% Introduction statistic on drug repurposing
\bibitem{DeNovo} Paul, Steven M., et al. "How to improve R\&D productivity: the pharmaceutical industry's grand challenge." Nature reviews Drug discovery 9.3 (2010): 203-214.

% Method 1 that uses molecular docking (got from Affinity2Vec citations)
\bibitem{Docking1} Alonso, Hernan, Andrey A. Bliznyuk, and Jill E. Gready. "Combining docking and molecular dynamic simulations in drug design." Medicinal research reviews 26.5 (2006): 531-568.

% Method 2 that uses molecular docking (got from Affinity2Vec citations)
\bibitem{Docking2} Kontoyianni, Maria. "Docking and virtual screening in drug discovery." Proteomics for drug discovery: Methods and protocols (2017): 255-266.

% Method 3 that uses molecular docking (got from Affinity2Vec citations)
\bibitem{Docking3} Mousavian, Zaynab, and Ali Masoudi-Nejad. "Drug–target interaction prediction via chemogenomic space: learning-based methods." Expert opinion on drug metabolism \& toxicology 10.9 (2014): 1273-1287.

\bibitem{alphafold}  “Highly accurate protein structure prediction with AlphaFold | Nature.” Accessed: Oct. 05, 2023. [Online]. Available: https://www.nature.com/articles/s41586-021-03819-2


% KronRLS
\bibitem{KronRLS} Pahikkala, Tapio, et al. "Toward more realistic drug–target interaction predictions." Briefings in bioinformatics 16.2 (2015): 325-337.

% SimBoost
\bibitem{SimBoost} He, Tong, et al. "SimBoost: a read-across approach for predicting drug–target binding affinities using gradient boosting machines." Journal of cheminformatics 9.1 (2017): 1-14.

% Modality-DTA
\bibitem{Modality} Yang, Xixi, et al. "Modality-dta: Multimodality fusion strategy for drug-target affinity prediction." IEEE/ACM Transactions on Computational Biology and Bioinformatics (2022).

% DeepDTA and CI metric
\bibitem{DeepDTA} Öztürk, Hakime, Arzucan Özgür, and Elif Ozkirimli. "DeepDTA: deep drug–target binding affinity prediction." Bioinformatics 34.17 (2018): i821-i829.

% GraphDTA
\bibitem{GraphDTA} Nguyen, Thin, et al. "GraphDTA: predicting drug–target binding affinity with graph neural networks." Bioinformatics 37.8 (2021): 1140-1147.

% GDGRU-DTA
\bibitem{GDGRU} Zhijian, Lyu, et al. "GDGRU-DTA: Predicting Drug-Target Binding Affinity Based on GNN and Double GRU." arXiv preprint arXiv:2204.11857 (2022).

% Affinity2Vec
\bibitem{Affinity2Vec} Thafar, Maha A., et al. "Affinity2Vec: drug-target binding affinity prediction through representation learning, graph mining, and machine learning." Scientific reports 12.1 (2022): 4751.

% DeepCPI
\bibitem{DeepCPI} Wan, Fangping, et al. "DeepCPI: a deep learning-based framework for large-scale in silico drug screening." Genomics, proteomics \& bioinformatics 17.5 (2019): 478-495.

% S2GC + SAGE
\bibitem{S2GC} Ma, Dong, Shuang Li, and Zhihua Chen. "Drug-target binding affinity prediction method based on a deep graph neural network." Mathematical Biosciences and Engineering 20.1 (2023): 269-282.

% SMT-DTA
\bibitem{SMT} Pei, Qizhi, et al. "SMT-DTA: Improving Drug-Target Affinity Prediction with Semi-supervised Multi-task Training." arXiv preprint arXiv:2206.09818 (2022).

% SEGSA-DTA
\bibitem{SEGSA} Gu, Yuliang, et al. "Protein–ligand binding affinity prediction with edge awareness and supervised attention." Iscience 26.1 (2023).

% iEdgeDTA
\bibitem{iEdge} Suviriyapaisal, Natchanon, and Duangdao Wichadakul. "iEdgeDTA: integrated edge information and 1D graph convolutional neural networks for binding affinity prediction." (2023).

% DeepGS
\bibitem{DeepGS} Lin, Xuan. "DeepGS: Deep representation learning of graphs and sequences for drug-target binding affinity prediction." arXiv preprint arXiv:2003.13902 (2020).

% Original KIBA dataset
\bibitem{KIBA} Tang, Jing, et al. "Making sense of large-scale kinase inhibitor bioactivity data sets: a comparative and integrative analysis." Journal of Chemical Information and Modeling 54.3 (2014): 735-743.

% iFeature
\bibitem{iFeature} Chen, Zhen, et al. "iFeature: a python package and web server for features extraction and selection from protein and peptide sequences." Bioinformatics 34.14 (2018): 2499-2502.

% RDKit
\bibitem{RDKit} Bento, A. Patrícia, et al. "An open source chemical structure curation pipeline using RDKit." Journal of Cheminformatics 12 (2020): 1-16.

% TDC
\bibitem{TDC} Huang, Kexin, et al. "Therapeutics data commons: Machine learning datasets and tasks for drug discovery and development." arXiv preprint arXiv:2102.09548 (2021).

% Smi2Vec representation method
\bibitem{Smi2Vec} Zhe Quan, Xuan Lin, Zhi-Jie Wang, Yan Liu, Fan Wang, and Kenli Li, ‘A system for learning atoms based on long short-term memory recurrent neural networks’, in 2018 IEEE International Conference on Bioinformatics and Biomedicine, pp. 728–733, (2018).

% Prot2Vec representation method
\bibitem{Prot2Vec} Ehsaneddin Asgari and Mohammad RK Mofrad, ‘Continuous distributed representation of biological sequences for deep proteomics and genomics’, PloS one, 10(11), e0141287, (2015).
    
% More papers if needed...

\end{thebibliography}
\end{document}